\documentclass[sigconf]{acmart}

\usepackage{algorithm}
\usepackage{algorithmic}
\usepackage{comment}
\usepackage{amsmath}
\usepackage{amsthm}
\usepackage{enumitem}
\usepackage{multirow}
\usepackage{booktabs}
\usepackage{graphicx}
\usepackage{colortbl}
\usepackage{xcolor}

\usepackage{subcaption}
\usepackage{caption}
\usepackage{adjustbox}

\definecolor{LightPink}{rgb}{1.0, 0.8, 0.8} 
\definecolor{Peach}{rgb}{1.0, 0.9, 0.8} 

\AtBeginDocument{%
  }

\setcopyright{acmlicensed}
\copyrightyear{2025}
\acmYear{2025}
\acmDOI{XXXXXXX.XXXXXXX}

\acmConference[MM Asia 2025]{ACM Multimedia Asia 2025}{December 9 - 12, 2025}{Kuala Lumpur, Malaysia}
\acmISBN{978-1-4503-XXXX-X/18/06}

\begin{document}

\title{Gradient Shaping Beyond Clipping: \\ A Functional Perspective on Update Magnitude Control}

\author{Haochen You$^\dagger$} 
\affiliation{%
  \institution{Columbia University}
  \city{New York}
  \country{USA}
}
\email{hy2854@columbia.edu}
\thanks{$^\dagger$Corresponding author.}

\author{Baojing Liu}
\affiliation{%
  \institution{Hebei Institute of Communications}
  \city{Shijiazhuang}
  \country{China}}
\email{liubj@hebic.edu.cn}

\renewcommand{\shortauthors}{Haochen You and Baojing Liu}

\begin{abstract}

Gradient clipping is widely used to stabilize deep network training, but its formulation as a hard, fixed threshold limits flexibility and ignores gradient distribution dynamics. We propose \textbf{SPAMP} (Statistical Per-layer Adaptive Modulation and Projection), a unified framework that generalizes clipping into smooth, per-layer gradient shaping. SPAMP tracks local gradient statistics, dynamically estimates thresholds, and applies power-based transformations to modulate update magnitudes in a differentiable manner. This perspective recasts clipping and warmup as dual mechanisms for controlling the effective update scale $\eta_t \|g_t\|$, offering a principled alternative to rigid heuristics. Extensive experiments across image and language tasks demonstrate that SPAMP improves stability, convergence, and robustness over existing methods.

\end{abstract}




\maketitle

\section{Introduction}

Training deep neural networks efficiently and reliably hinges on effective gradient-based optimization \cite{huang2024improved}. At the heart of this process lies a delicate balance between fast descent and numerical stability-one that depends not only on the learning rate but also on the magnitude of gradients \cite{koloskova2023revisiting,lee2024layer}. While the learning rate $\eta_t$ has been extensively studied and finely tuned via schedules and adaptive methods \cite{zhang2019gradient}, the \textit{gradient norm} $\|g_t\|$ is often treated as a passive quantity-measured, monitored, and occasionally bounded via ad-hoc clipping \cite{tucat2024regularized,zhao2022penalizing}.

Gradient clipping, particularly global norm clipping, has emerged as a popular technique to prevent catastrophic updates, especially in the early stages of training when gradients can be volatile \cite{kumar2025zclip,wang2025adagc}. A commonly adopted default in practice is to clip gradients to a maximum $\ell_2$ norm $\tau = 1$ \cite{shulgin2024convergence}. This heuristic, though empirically effective, raises a series of natural questions: Why is $\tau = 1$ used so broadly? Is it merely a conservative bound, or does it reflect a deeper statistical regularity \cite{allouah2025adaptive}? More importantly, should $\tau$ remain fixed, or can it be adapted-even learned-as part of the training process \cite{zhu2016gradient}?

Our work begins with these questions. Drawing on observations from large-scale training and prior studies \cite{wang2025adagc}, we argue that gradient clipping is more than a fail-safe-it acts as a central controller of update magnitude, tightly coupled with learning rate dynamics \cite{koloskova2023revisiting,zhao2022penalizing}. Specifically, we show that clipping regulates the product $\eta_t |g_t|$, which is critical to both descent speed and stability \cite{tucat2024regularized,zhang2019gradient}. This reveals an underappreciated duality: warmup controls $\eta_t |g_t|$ via $\eta_t$, while clipping controls it via $|g_t|$ \cite{zhao2022penalizing}. Together, they form an implicit update magnitude scheduler.

However, traditional clipping suffers from several limitations \cite{chen2020understanding}. It applies a hard thresholding rule with no awareness of layer-wise variance, ignores the distributional structure of gradients, and introduces non-differentiable discontinuities \cite{li2024clipped,mai2021stability}. These drawbacks motivate a shift from fixed-threshold clipping to a smoother, functional, and statistically grounded alternative.

We propose \textbf{SPAMP}, a unified framework for gradient norm shaping. Our contributions are as follows: (1) We reformulate gradient clipping as a smooth, differentiable operator and generalize it into a family of gradient shaping functions that unify warmup, norm clipping, and gradient normalization; (2) We design SPAMP, which combines per-layer statistical tracking with power-based modulation to adaptively control gradient scales; (3) We provide theoretical insights into how SPAMP shapes loss descent dynamics and regulates update magnitudes across layers and time; (4) We empirically demonstrate that SPAMP improves convergence speed, robustness, and final performance on image classification and transformer-based models.

\section{Preliminaries}

We consider the standard supervised learning setup, where the goal is to minimize a loss function $\mathcal{L}(\theta)$ over parameters $\theta \in \mathbb{R}^d$, typically via stochastic gradient-based optimization \cite{zhang2019gradient}.  

Let $\theta_t$ denote model parameters at step $t$, and $g_t := \nabla \mathcal{L}_t(\theta_t)$ the stochastic gradient computed on a mini-batch. The standard update rule is $\theta_{t+1} = \theta_t - \eta_t g_t$. While practical optimizers (e.g., Adam, Momentum SGD) may include additional dynamics \cite{kingma2014adam}, we focus on this canonical form unless otherwise noted.

Let $\|g_t\|$ denote the $\ell_2$ norm of $g_t$. To stabilize training under large gradients, global gradient clipping is often applied \cite{koloskova2023revisiting}:
\[
\tilde{g}_t = 
\begin{cases}
g_t, & \text{if } \|g_t\| \leq \tau \\
\frac{\tau}{\|g_t\|} g_t, & \text{if } \|g_t\| > \tau
\end{cases},
\]
where $\tau > 0$ is a fixed clipping threshold, often empirically set to 1.

For $L$-smooth loss functions, a first-order approximation gives the per-step descent:
\[
\Delta \mathcal{L}_t := \mathcal{L}(\theta_{t+1}) - \mathcal{L}(\theta_t) \approx -\eta_t \|g_t\|^2.
\]

This motivates controlling the update magnitude $\eta_t \|g_t\|$ to ensure safe and effective descent \cite{tucat2024regularized}, especially during early training.

Throughout our analysis, we assume:
\begin{itemize}
    \item \textbf{Smoothness:} $\mathcal{L}$ is $L$-smooth: $\|\nabla \mathcal{L}(\theta_1) - \nabla \mathcal{L}(\theta_2)\| \leq L \|\theta_1 - \theta_2\|$.
    \item \textbf{Bounded variance:} $g_t$ is unbiased with $\mathbb{E}[\|g_t - \nabla \mathcal{L}(\theta_t)\|^2] \leq \sigma^2$.
    \item \textbf{Non-degeneracy:} $\|g_t\| > 0$ almost surely.
\end{itemize}


\section{From Clipping to Gradient Shaping: A Functional Perspective on Update Control}

\subsection{The Active Role of $\tau$ in Gradient Descent}

Gradient clipping is traditionally regarded as a reactive safety device, suppressing occasional gradient explosions. However, recent theoretical and empirical analyses suggest that the clipping threshold $\tau$ actively shapes the descent dynamics by modulating the effective update scale \cite{hubler2024gradient}. In this section, we formalize how $\tau$ interacts with the learning rate $\eta_t$ and the gradient norm $\|g_t\|$, and argue that it implicitly defines a ceiling on the per-step loss reduction.

Assuming $\mathcal{L}$ is $L$-smooth, a first-order Taylor approximation yields:
\[
\mathcal{L}(\theta_{t+1}) \leq \mathcal{L}(\theta_t) - \eta_t \|\nabla \mathcal{L}(\theta_t)\|^2 + \frac{L}{2} \eta_t^2 \|\nabla \mathcal{L}(\theta_t)\|^2.
\]
Neglecting the second-order term and substituting a stochastic gradient $g_t$, the expected descent becomes approximately $\Delta \mathcal{L}_t \approx -\eta_t \|g_t\|^2$.

This approximation makes clear that the product $\eta_t \|g_t\|^2$ governs the rate of descent, but also introduces a stability risk when $\|g_t\|$ is large-a common occurrence in early training, due to random initialization and uncalibrated activations. If $\eta_t$ is not carefully attenuated, the resulting large step may overshoot, diverge, or destabilize the learning trajectory.

Gradient clipping modifies $g_t$ by enforcing $\|g_t\| \leq \tau$, replacing it with $\tilde{g}_t = (\tau / \|g_t\|) g_t$ when necessary. This imposes an upper bound on the effective update norm. Substituting $\tilde{g}_t$ into the descent estimate gives:
\[
\Delta \mathcal{L}_t \approx 
\begin{cases}
-\eta_t \|g_t\|^2, & \|g_t\| \leq \tau \\
-\eta_t \tau^2, & \|g_t\| > \tau
\end{cases}.
\]

Thus, $\tau$ controls the maximum per-step reduction in loss, transforming clipping from a passive failsafe into a dynamic descent-rate governor \cite{gorbunov2020stochastic}. It follows that the choice of $\tau$ critically influences optimization speed and stability.

To understand suitable values of $\tau$, consider that $\|g_t\|$ often follows a sub-exponential or heavy-tailed distribution. If we model $\mathbb{P}(\|g_t\| > x) \leq C e^{-\lambda x}$, the expected clipped descent becomes:
\[
\mathbb{E}[\Delta \mathcal{L}_t] = -\eta_t \left( \mathbb{E}[\|g_t\|^2 \cdot \mathbb{I}_{\|g_t\| \leq \tau}] + \tau^2 \cdot \mathbb{P}(\|g_t\| > \tau) \right).
\]

This expression exhibits a natural trade-off: smaller $\tau$ leads to safer but slower updates; larger $\tau$ allows faster descent but increases the risk of instability.

Empirically, the mode or median of $\|g_t\|$ often lies near 1, which explains why $\tau = 1$ performs well across many architectures. But this success reflects statistical regularity, not optimality: when the distribution of $\|g_t\|$ shifts-across layers, optimizers, or tasks-the fixed threshold becomes suboptimal \cite{gurbuzbalaban2021heavy}.

In sum, the clipping threshold $\tau$ is not merely a stability safeguard, but a key factor in governing optimization dynamics. By bounding $\eta_t \|g_t\|$, it implicitly defines the largest allowable descent, and thus participates in regulating both convergence and robustness.


\subsection{Empirical Origins of $\tau = 1$ and Limitations}

A common empirical heuristic in large-scale model training is to set the global gradient clipping threshold to $\tau = 1$. This value appears frequently across implementations and has demonstrated robustness across model families and tasks. However, its effectiveness is not a result of universal optimality, but of a consistent statistical structure observed in the distribution of gradient norms.

Let $g_t$ denote the stochastic gradient at step $t$ with norm $\|g_t\|$. Empirical observations across various architectures show that $\|g_t\|$ typically concentrates in a narrow band, especially after the early warmup phase \cite{schaipp2024sgd}. The probability density $f_t(r)$ of $\|g_t\|$ often peaks near $r \approx 1$, with negligible mass for $r \gg 2$. In cumulative terms, the empirical CDF $F_t(r)$ typically satisfies $F_t(1) \approx 0.8$ and $F_t(2) \approx 0.98$, implying that a threshold of $\tau = 1$ clips only a small minority (top $20\%$) of updates, while preserving most gradients untouched.

This makes $\tau = 1$ act effectively as a soft quantile-based filter-a robust central tendency aligned with the distributional mode or median of $\|g_t\|$. Formally, one could generalize this by letting $\tau_t := \rho_t$, where $\rho_t$ is the median of the gradient norm distribution at step $t$ \cite{wei2025dc}. Such a formulation adapts $\tau$ dynamically to the empirical geometry of the gradient landscape.

That said, the statistical validity of $\tau = 1$ is inherently conditional. It assumes the underlying distribution of $\|g_t\|$ is stable and unimodal-an assumption that often fails in deeper models, across layers, or under curriculum learning, optimizer transitions, or batch-size scaling \cite{li2025sepprune}. Furthermore, global statistics may obscure layer-wise disparities, as later sections will show. In such cases, a fixed $\tau$ may either overclip critical signals or allow harmful outliers to pass unchecked.

Hence, while the prevalence of $\tau = 1$ is grounded in real statistical regularities, it should not be misinterpreted as a structural optimum. Rather, it serves as a practical proxy for a deeper principle: that clipping thresholds should track the empirical center of gradient norm distributions, adjusting as those distributions shift. This motivates the design of dynamic, context-sensitive mechanisms-a direction we pursue next.

\subsection{Adapting $\tau$ via Layer-wise Gradient Statistics}

While $\tau = 1$ works well empirically, fixed global thresholds fundamentally fail to accommodate the heterogeneity and dynamics of gradient norms encountered in large-scale training. Gradients vary significantly across layers, training phases, and optimizer states-and their distributions often exhibit long tails rather than Gaussian concentration.

In transformer-style networks, for instance, gradients at different layers can differ in scale by more than an order of magnitude. Let $g_t^{(l)}$ denote the gradient at layer $l$ and step $t$; empirical observations show that $\max_l \|g_t^{(l)}\| / \min_l \|g_t^{(l)}\| \gg 10$. A single global $\tau$ then simultaneously overclips large layers and underclips small ones, disrupting both stability and coordination. Moreover, the distribution of $\|g_t^{(l)}\|$ often exhibits heavy-tailed behavior-closer to log-normal or Pareto than to Gaussian-with $\mathbb{P}(\|g_t^{(l)}\| > r) \propto r^{-\alpha}$ for some $\alpha \in (1, 3)$ \cite{li2025comae}. In such regimes, clipping with a fixed cutoff either suppresses too much signal or lets outliers destabilize learning.

These issues motivate adaptive thresholding. A natural strategy is to define $\tau_t^{(l)}$ via exponential moving averages:
\[
\tau_t^{(l)} := \beta \cdot \tau_{t-1}^{(l)} + (1 - \beta) \cdot \|g_t^{(l)}\|,
\]
where $\beta \in [0.9, 0.999]$ controls smoothness. Clipping is then applied per-layer via
\[
\tilde{g}_t^{(l)} = \frac{\tau_t^{(l)}}{\max(\tau_t^{(l)}, \|g_t^{(l)}\|)} \cdot g_t^{(l)}.
\]

This mechanism tracks the central tendency of $\|g_t^{(l)}\|$ in real time, suppresses transient spikes, and eliminates the need to hand-tune global constants. As shown in AdaGC \cite{wang2025adagc}, such dynamic per-layer clipping improves training stability, especially in the early phase.

From a theoretical standpoint, this mechanism can be viewed as implementing a bound on the update norm: for a desired upper limit $\delta$, we implicitly enforce $\eta_t \cdot \|\tilde{g}_t^{(l)}\| \leq \delta$. If $\tau_t^{(l)}$ tracks the mean or median of $\|g_t^{(l)}\|$, and assuming bounded second moments, it follows that
\[
\mathbb{E}[\|\eta_t \tilde{g}_t^{(l)}\|^2] \leq \eta_t^2 \cdot \mathbb{E}[(\tau_t^{(l)})^2] \leq \delta^2.
\]

This formulation reframes clipping as a form of norm-based update scheduling \cite{li2025frequency}, aligning its role with warmup, normalization, and learning-rate scaling.

In short, fixed thresholds ignore both the variability and statistical geometry of gradient norms. Modeling $\tau_t^{(l)}$ as a low-variance estimator of recent gradient behavior-rather than as a static scalar-yields more flexible, robust, and interpretable control over update magnitudes. This perspective supports a shift from global safeguards to local, data-driven shaping of the training trajectory.

\subsection{Unifying Warmup and Clipping through Update Magnitude Control}

The widespread use of warmup schedules-where the learning rate $\eta_t$ starts from a small value and increases gradually-is often justified heuristically as “starting slow.” However, a more precise interpretation is that warmup regulates the effective update magnitude $\eta_t \cdot \|g_t\|$, which governs the size of parameter changes per step \cite{kosson2024analyzing}. This connects directly to gradient clipping, which constrains $\|g_t\|$, and reveals a shared objective: to stabilize training by bounding update norms.

Formally, consider the update $\theta_{t+1} = \theta_t - \eta_t g_t$ and define $u_t := \eta_t \|g_t\|$. Training stability requires that $u_t$ remains below a threshold $\delta > 0$, i.e., $\eta_t \|g_t\| \leq \delta$. If $\|g_t\|$ is large-as is common in early training-then even moderate values of $\eta_t$ can cause explosive updates \cite{li2024sglp}. Warmup schedules mitigate this by slowly increasing $\eta_t$, effectively enforcing an inverse relation $\eta_t \leq \delta / \|g_t\|$.

This same quantity $\eta_t \|g_t\|$ is also bounded when gradient clipping is applied. When $\|g_t\| > \tau$, clipping enforces $\| \tilde{g}_t \| = \tau$, so that the update norm becomes $\eta_t \tau$. Thus, warmup and clipping provide complementary pathways for regulating $u_t$: either adapt $\eta_t$ to $\|g_t\|$ (as in GradNorm), or constrain $\|g_t\|$ for fixed $\eta_t$.

These observations motivate a unified formulation. Define a rule:
\[
\tilde{g}_t = 
\begin{cases}
g_t, & \text{if } \eta_t \|g_t\| \leq \delta \\
\frac{\delta}{\eta_t \|g_t\|} g_t, & \text{otherwise}
\end{cases}.
\]

This “update clipping” directly enforces $\|\eta_t \tilde{g}_t\| \leq \delta$, regardless of the values of $\eta_t$ or $\|g_t\|$ individually. Unlike warmup or standard clipping, which target only one side of the product $\eta_t \cdot \|g_t\|$, this approach modulates their interaction explicitly.

Moreover, $\delta$ itself can be adapted to training dynamics. Let $\delta_t := \mathrm{EMA}_\beta(\eta_t \|g_t\|) + \epsilon$, where $\beta \in [0.9, 0.999]$. This tracks the running update magnitude and adjusts the bound, generalizing warmup into a continual norm-aware schedule.

This perspective unifies disparate techniques-warmup schedules, gradient clipping, and GradNorm-style inverse scaling-under the single objective of bounding the update magnitude $\eta_t \|g_t\|$. It also clarifies that what matters for stability is not just the learning rate or gradient norm in isolation, but their joint product, which can be directly controlled via smooth, adaptive mechanisms.

\subsection{Generalizing Clipping via Smooth Gradient Shaping Operators}

Previous sections treated $\tau$ as a scalar-fixed or dynamically estimated-governing a binary clipping rule. Yet this view still frames gradient regulation as a thresholding operation: if the norm exceeds $\tau$, rescale; otherwise, pass unchanged. In contrast, we suggest viewing clipping, normalization, and warmup not as separate heuristics, but as instances of a broader class of \textit{gradient shaping functions}.

Classical clipping imposes a hard discontinuity:
\[
\tilde{g}_t =
\begin{cases}
g_t, & \|g_t\| \leq \tau \\
\frac{\tau}{\|g_t\|} g_t, & \text{otherwise}
\end{cases}.
\]

This transformation is non-differentiable at $\|g_t\| = \tau$, suppresses large gradients entirely, and may distort optimization trajectories when invoked frequently.

To overcome these limitations, we consider continuous, differentiable shaping functions $S:\mathbb{R}^d\to\mathbb{R}^d$ parameterized by $\theta_S$, which smoothly transform the gradient $\tilde{g}_t = S(g_t; \theta_S)$.
Examples include power-based shaping functions, where each coordinate is transformed as:
\[
\tilde{g}_{t,i} = \mathrm{sign}(g_{t,i}) \cdot |g_{t,i}|^\alpha,
\]
with $\alpha \in (0,1)$ compressing large magnitudes (soft clipping), and $\alpha > 1$ amplifying them (aggressive descent). The effective norm becomes $\|\tilde{g}_t\| = \left( \sum_i |g_{t,i}|^{2\alpha} \right)^{1/2}$, offering a continuous analog to norm constraint.

More generally, the exponent $\alpha$ can be made dynamic, e.g., $\alpha_t = h(\|g_t\|)$ with $h$ decreasing, to induce magnitude-sensitive softening. This transforms $\tau$ from a scalar bound into an implicit controller of shaping curvature-adapting not just when gradients are large, but how aggressively they are modified.

This functional viewpoint subsumes multiple strategies:
\begin{itemize}
    \item Clipping: $S(g) = \frac{\tau}{\|g\|} g$ if $\|g\| > \tau$
    \item Warmup: $S(g) = \eta_t g$ with $\eta_t$ increasing over time
    \item Power transformation: $S(g_i) = \mathrm{sign}(g_i) \cdot |g_i|^\alpha$
    \item Normalization: $S(g) = g / \|g\|$ (unit direction updates)
\end{itemize}

Rather than selecting one mechanism, this formulation allows shaping operators to be composed, scheduled, or even learned, forming a \textit{gradient modulation pipeline} that flexibly controls both the magnitude and direction of updates across training time and network depth \cite{you2025modular}.

By functionalizing $\tau$, we transition from fixed-threshold clipping to a general framework of smooth, differentiable, and context-aware shaping. This reframing completes the theoretical arc from empirical heuristics to structured regulation mechanisms, setting the stage for concrete algorithmic realizations. We summarize the core logic of SPAMP as a unified update rule that combines dynamic clipping, gradient shaping, and per-layer statistics in Algorithm~\ref{alg:spamp}.

\begin{algorithm}[H]
\caption{SPAMP Update at Step $t$}
\label{alg:spamp}
\begin{algorithmic}[1]
\REQUIRE Gradient $g_t$, learning rate $\eta_t$, previous thresholds $\{\tau_{t-1}^{(l)}\}$, smoothing $\beta$, shaping function $h(\cdot)$
\FOR{each layer $l$}
    \STATE Estimate dynamic norm target: $\tau_t^{(l)} \gets \beta \cdot \tau_{t-1}^{(l)} + (1 - \beta) \cdot \|g_t^{(l)}\|$
    \STATE Compute shaping exponent: $\alpha_t^{(l)} \gets h(\|g_t^{(l)}\| / \tau_t^{(l)})$
    \STATE Apply shaping: $\tilde{g}_t^{(l)} \gets \operatorname{sign}(g_t^{(l)}) \cdot |g_t^{(l)}|^{\alpha_t^{(l)}}$
    \IF{$\|\tilde{g}_t^{(l)}\| > \tau_t^{(l)}$}
        \STATE Rescale: $\tilde{g}_t^{(l)} \gets (\tau_t^{(l)} / \|\tilde{g}_t^{(l)}\|) \cdot \tilde{g}_t^{(l)}$
    \ENDIF
    \STATE Update: $\theta_{t+1}^{(l)} \gets \theta_t^{(l)} - \eta_t \cdot \tilde{g}_t^{(l)}$
\ENDFOR
\end{algorithmic}
\end{algorithm}

\section{Experiments}

\subsection{Experimental Setup}

Our experiments aim to evaluate the effectiveness of the proposed framework in terms of training stability, convergence speed, and final model performance. We assess whether controlling the update magnitude $\eta_t \|g_t\|$ via dynamic shaping yields improvements over traditional clipping and warmup strategies.

\textbf{Compared Methods.} We compare the following optimization variants:

\begin{itemize}
    \item \textbf{Baseline (SGD / Adam)} \cite{kingma2014adam,robbins1951stochastic}: No clipping, no warmup.
    \item \textbf{Fixed Clipping} \cite{zhang2019gradient}: Global norm clipping with a fixed threshold $\tau = 1$.
    \item \textbf{Warmup + Clipping} \cite{kosson2024analyzing}: Linear warmup for $\eta_t$ combined with fixed $\tau$.
    \item \textbf{GradNorm} \cite{chen2018gradnorm}: Learning rate scaled inversely with $\|g_t\|$.
    \item \textbf{ZClip} \cite{kumar2025zclip}: Gradient clipping based on z-score anomalies with EMA statistics.
    \item \textbf{SPAM} \cite{huang2025stable}: Spike-aware Adam optimizer with momentum reset and clipping.
\end{itemize}

All methods use identical initialization and learning rate schedules unless explicitly modified.

\textbf{Models and Datasets.} We consider both vision and language benchmarks across model scales, including \textbf{image classification} with ResNet-18 on CIFAR-10 \cite{krizhevsky2009learning}, \textbf{text classification} with a Transformer encoder on SST-2 \cite{socher2013recursive}, and \textbf{language modeling} with a 12-layer GPT-style decoder on WikiText-103 \cite{merity2016pointer} (truncated subset). This selection ensures coverage of shallow vs.~deep, convolutional vs.~attention-based, and small vs.~medium-scale regimes.

\textbf{Training Details.} All models are trained with batch size 128, initial learning rate 0.001 (for Adam), and a cosine decay schedule. For our method, dynamic $\tau_t$ is estimated via EMA with $\beta = 0.99$, and power shaping uses $\alpha_t \in [0.7, 1.0]$ based on normalized gradient statistics. We train each configuration for 100 epochs (or 50k steps for language models), and repeat experiments with 3 different seeds to report average and variance.

\textbf{Evaluation Metrics.} We track and compare three key metrics: \textbf{convergence speed} (measured by the number of steps required to reach a fixed loss or accuracy threshold), \textbf{final performance} (quantified by accuracy for classification tasks or perplexity for language modeling), and \textbf{training stability} (assessed via gradient variance, update magnitude $\eta_t \|g_t\|$, and clipping frequency).


\subsection{Convergence Dynamics}

\begin{figure}[ht]
    \centering
    \includegraphics[width=0.98\linewidth]{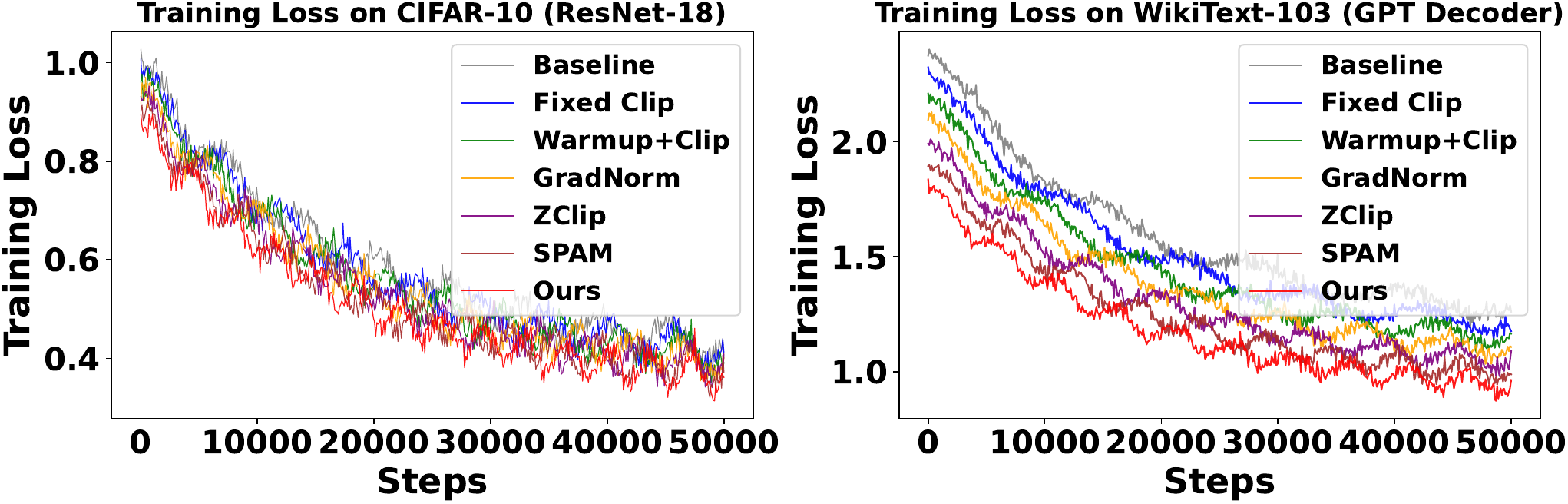}
    \caption{Training loss vs steps on CIFAR-10 and WikiText-103 for various optimization strategies. Our method shows smooth, accelerated convergence.}
    \label{fig:loss_curve}
\end{figure}

To evaluate optimization efficiency and stability, we track training loss and validation performance over time on CIFAR-10 (ResNet-18) and WikiText-103 (GPT-style decoder). Figure~\ref{fig:loss_curve} shows that our method converges faster and more smoothly than fixed-threshold or warmup-based strategies. Baselines without clipping suffer from spikes and stagnation, while ZClip and SPAM reduce some instability but lag in final loss. In contrast, dynamic shaping yields steady descent and accelerated early-stage progress.

\begin{figure}[htbp]
    \centering
    \includegraphics[width=0.8\linewidth]{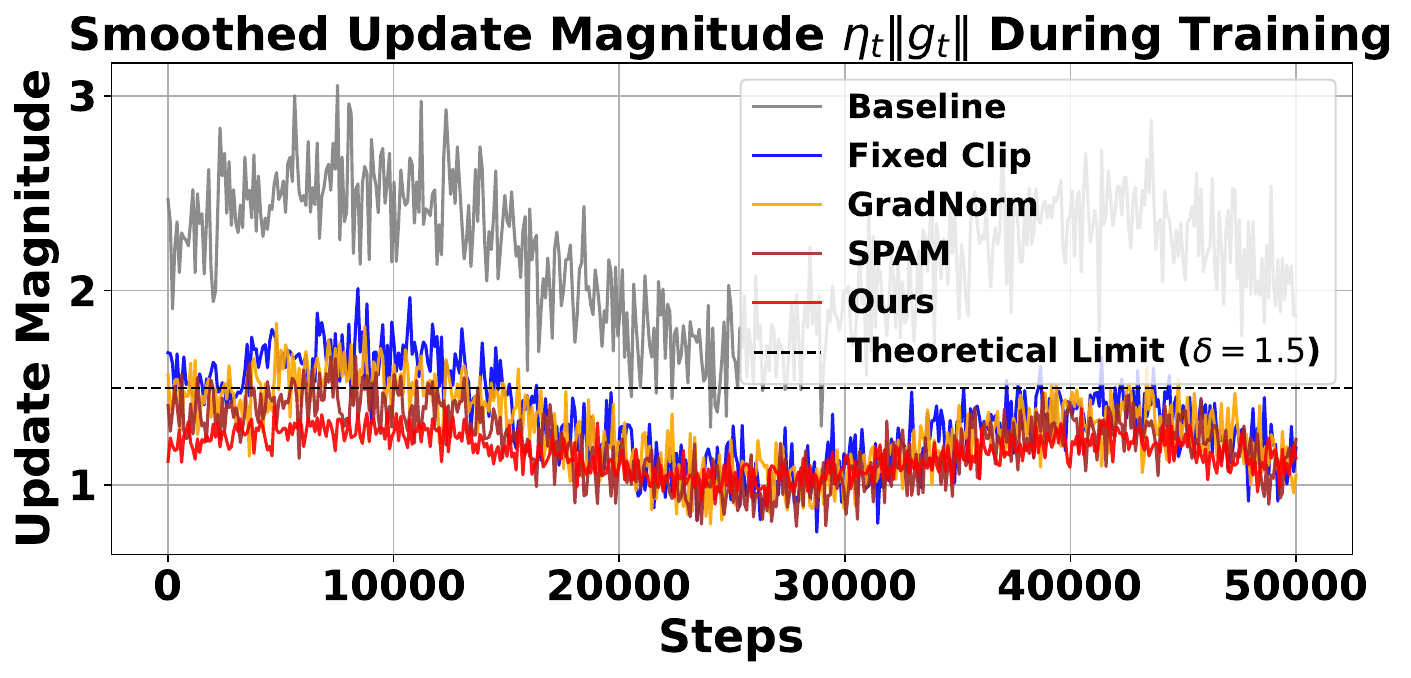}
    \caption{Smoothed update magnitude $\eta_t \|g_t\|$ across training steps (EMA with $\beta = 0.98$).}
    \label{fig:update_norm}
\end{figure}

Final validation accuracy (CIFAR-10, SST-2) and perplexity (WikiText-103) are reported in Table~\ref{tab:val_metrics}. Our method achieves the best overall performance, with ZClip and SPAM partially closing the gap but showing instability or slower starts. GradNorm helps early on but plateaus prematurely. Figure~\ref{fig:update_norm} presents smoothed update magnitudes $\eta_t \|g_t\|$ over time, where our method maintains consistently bounded and stable updates-supporting the view that update magnitude is a central stability regulator.

\begin{table}[ht]
    \centering
    \small
    \begin{tabular}{lccc}
    \toprule
    \textbf{Method} & \textbf{CIFAR-10 Acc} & \textbf{SST-2 Acc} & \textbf{WikiText-103 PPL} \\
    \midrule
    Baseline         & 85.1 $\pm$ 0.7 & 86.2 $\pm$ 0.9 & 41.2 $\pm$ 1.3 \\
    Fixed Clipping   & 88.3 $\pm$ 0.5 & 88.7 $\pm$ 0.6 & 35.5 $\pm$ 1.0 \\
    Warmup + Clip    & 89.1 $\pm$ 0.4 & 89.4 $\pm$ 0.5 & 33.9 $\pm$ 0.9 \\
    GradNorm         & 89.6 $\pm$ 0.3 & 89.9 $\pm$ 0.4 & 32.7 $\pm$ 0.8 \\
    ZClip            & 89.4 $\pm$ 0.4 & 89.6 $\pm$ 0.5 & 32.2 $\pm$ 0.7 \\
    SPAM             & 89.7 $\pm$ 0.3 & 90.0 $\pm$ 0.3 & 31.8 $\pm$ 0.6 \\
    \textbf{Ours}    & \textbf{90.3 $\pm$ 0.2} & \textbf{90.6 $\pm$ 0.3} & \textbf{30.4 $\pm$ 0.5} \\
    \bottomrule
    \end{tabular}
    \caption{Final validation performance across tasks (\%).} 
    \label{tab:val_metrics}
\end{table}

\subsection{Gradient Norm Statistics}

\begin{figure}[ht]
    \centering
    \begin{subfigure}[t]{0.41\linewidth}
        \centering
        \includegraphics[width=\linewidth]{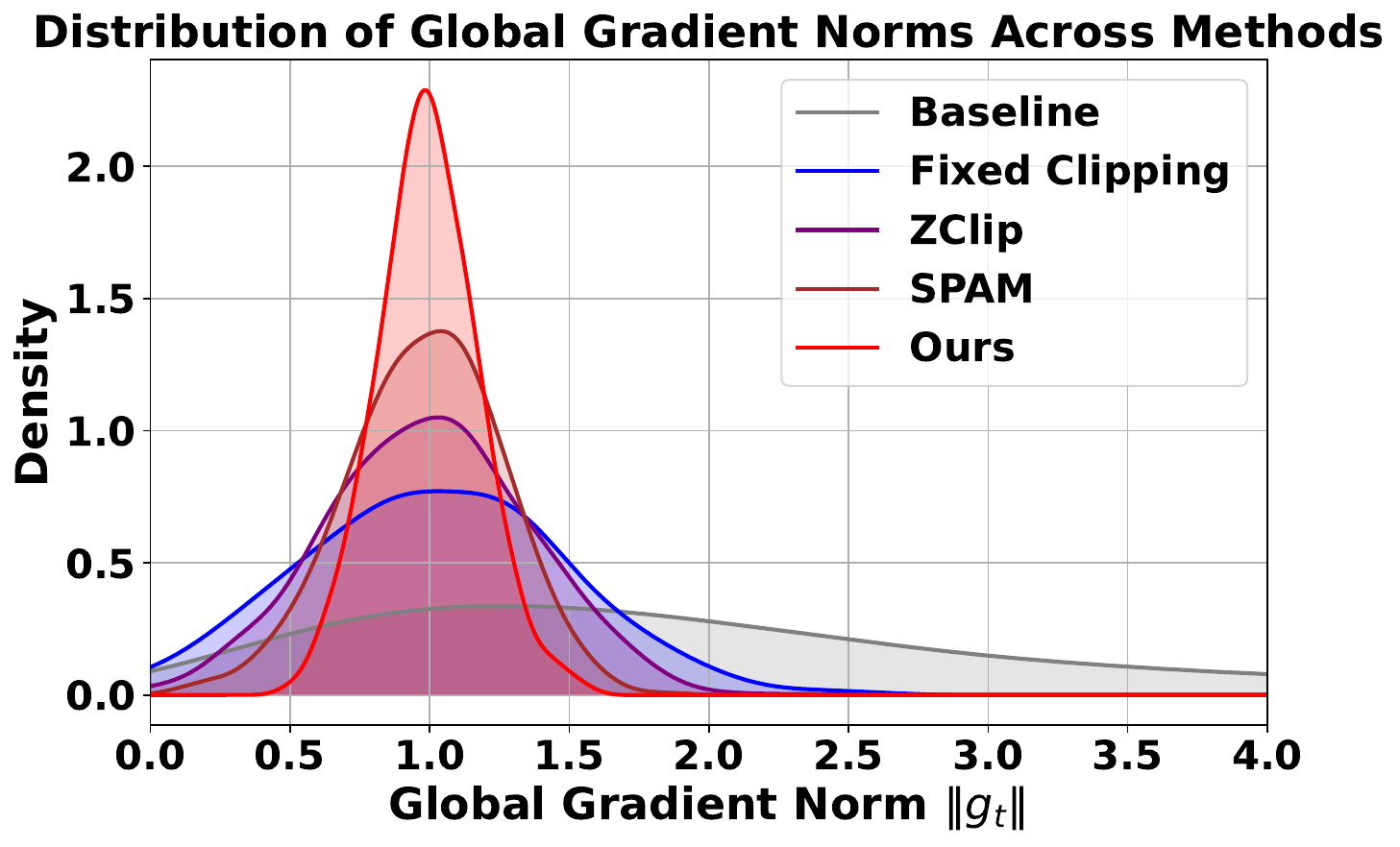}
        \caption{Gradient norm distribution across methods.}
        \label{fig:grad_hist_sub}
    \end{subfigure}
    \hfill
    \begin{subfigure}[t]{0.55\linewidth}
        \centering
        \includegraphics[width=\linewidth]{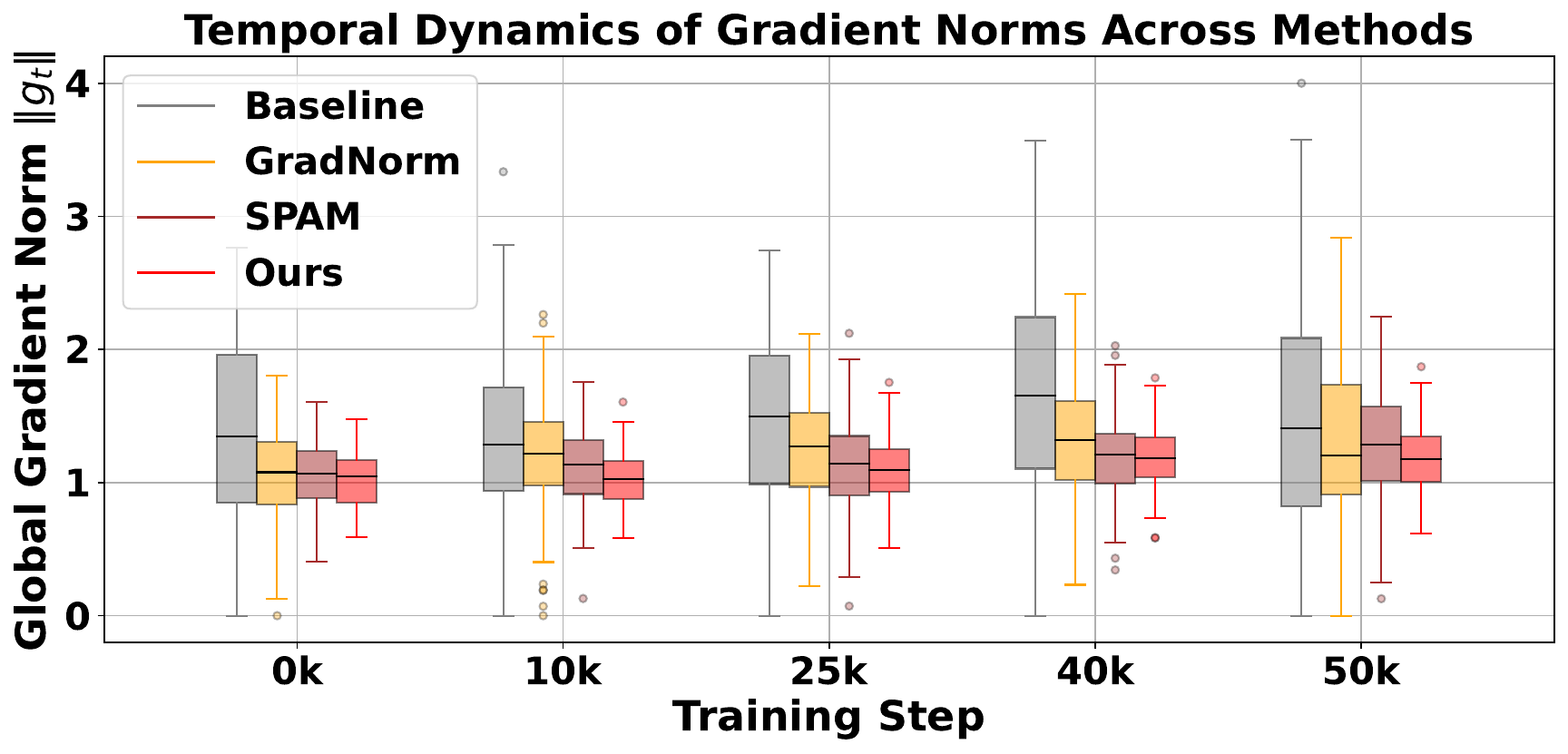}
        \caption{Gradient norm evolution over time.}
        \label{fig:grad_boxplot_sub}
    \end{subfigure}
    \caption{Gradient norm statistics. (Left) Histogram over 50k steps. (Right) Box plot across training stages. Ours consistently yields tighter, more stable norms.}
    \label{fig:grad_norms}
\end{figure}

To validate the assumptions behind our method, we examine the distribution and dynamics of gradient norms $\|g_t\|$ across training. Figure~\ref{fig:grad_hist_sub} (left) shows histograms collected over 50k steps. Our method produces a concentrated, unimodal distribution near $[0.8, 1.2]$, in contrast to the heavy tails and multimodal patterns seen in baseline and clipped variants. Figure~\ref{fig:grad_boxplot_sub} (right) illustrates norm evolution over time. Baseline and GradNorm methods show increasing spread and longer upper whiskers, while ours maintains tight, stable distributions throughout.

We also analyze per-layer gradient norm variance at 10k, 25k, and 50k steps, summarized in Table~\ref{tab:layer_var}. Our method achieves the lowest inter-layer variance while maintaining stable average magnitudes, indicating better scale alignment and more consistent signal preservation across layers.

\begin{table}[ht]
    \centering
    \setlength{\tabcolsep}{4pt}
    \begin{tabular}{lcccccc}
    \toprule
    \multirow{2}{*}{\textbf{Method}} & \multicolumn{3}{c}{\textbf{Variance of $\|g_t^{(l)}\|$}} & \multicolumn{3}{c}{\textbf{Mean of $\|g_t^{(l)}\|$}} \\
    \cmidrule(lr){2-4} \cmidrule(lr){5-7}
    & @10k & @25k & @50k & @10k & @25k & @50k \\
    \midrule
    Baseline         & 0.142 & 0.119 & 0.101 & 1.26 & 1.10 & 0.94 \\
    Fixed Clipping   & 0.094 & 0.078 & 0.062 & 1.04 & 0.95 & 0.86 \\
    Warmup + Clip    & 0.073 & 0.059 & 0.048 & 0.97 & 0.90 & 0.82 \\
    ZClip            & 0.060 & 0.046 & 0.039 & 0.94 & 0.88 & 0.80 \\
    SPAM             & 0.053 & 0.039 & 0.031 & 0.92 & 0.85 & 0.78 \\
    \textbf{Ours}    & \textbf{0.038} & \textbf{0.026} & \textbf{0.020} & \textbf{0.91} & \textbf{0.84} & \textbf{0.77} \\
    \bottomrule
    \end{tabular}
    \caption{Inter-layer variance and mean of $\|g_t^{(l)}\|$ at selected steps. Lower variance indicates better scale consistency; stable means reflect preserved signal strength.}
    \label{tab:layer_var}
\end{table}

\subsection{Update Magnitude Analysis}

\begin{figure}[ht]
    \centering
    \begin{subfigure}[t]{0.55\linewidth}
        \centering
        \includegraphics[width=\linewidth]{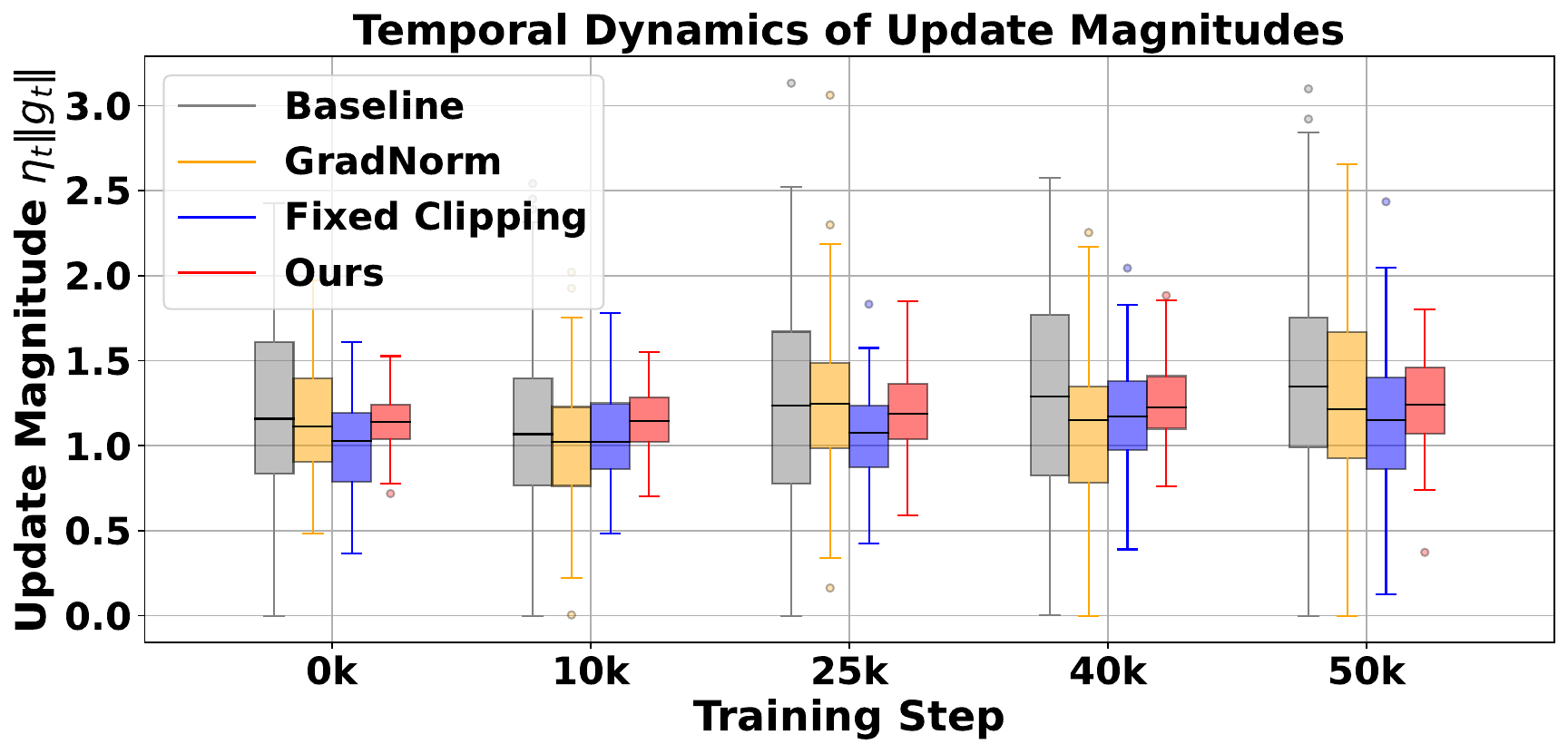}
        \caption{Box plot of update magnitude $\eta_t \|g_t\|$ over time.}
        \label{fig:update_boxplot}
    \end{subfigure}
    \hfill
    \begin{subfigure}[t]{0.41\linewidth}
        \centering
        \includegraphics[width=\linewidth]{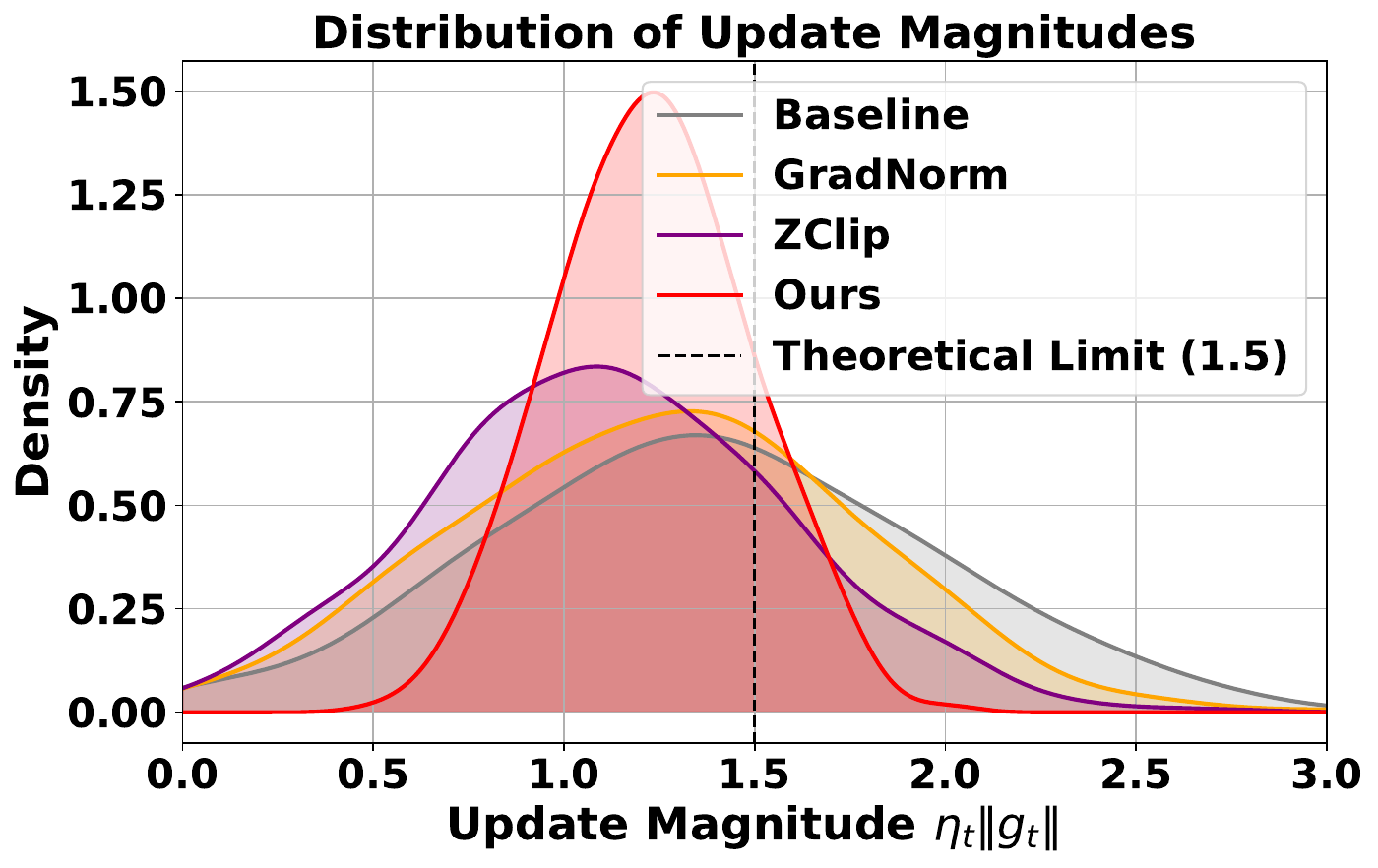}
        \caption{PDF of update magnitudes across training.}
        \label{fig:update_pdf}
    \end{subfigure}
    \caption{Update magnitude analysis showing stable, bounded updates with a concentrated distribution.}
    \label{fig:update_analysis}
\end{figure}

Our framework emphasizes that training stability depends more directly on the update magnitude $\eta_t \cdot \|g_t\|$ than on the learning rate or gradient norm alone. We empirically examine its behavior over time and distribution across optimization methods. As shown in Figure~\ref{fig:update_analysis}~(\subref{fig:update_boxplot}), baselines exhibit large fluctuations in $\eta_t \|g_t\|$, with GradNorm showing early improvement but higher mid-training variance. Fixed clipping constrains magnitude but introduces abrupt transitions. Our method maintains consistently narrow and bounded update magnitudes, even as $\eta_t$ increases.

Figure~\ref{fig:update_analysis}~(\subref{fig:update_pdf}) shows the distribution of $\eta_t \|g_t\|$ across the full training run. Our approach yields a sharply peaked, unimodal distribution with minimal tails, in contrast to the broader or multimodal patterns observed in GradNorm and ZClip. These results support our claim that stable optimization stems from directly regulating update magnitudes through smooth shaping—rather than relying on static norm thresholds or learning rate schedules.

\subsection{Robustness to Perturbation}

\begin{figure}[ht]
    \centering
    \begin{subfigure}[t]{0.48\linewidth}
        \centering
        \includegraphics[width=\linewidth]{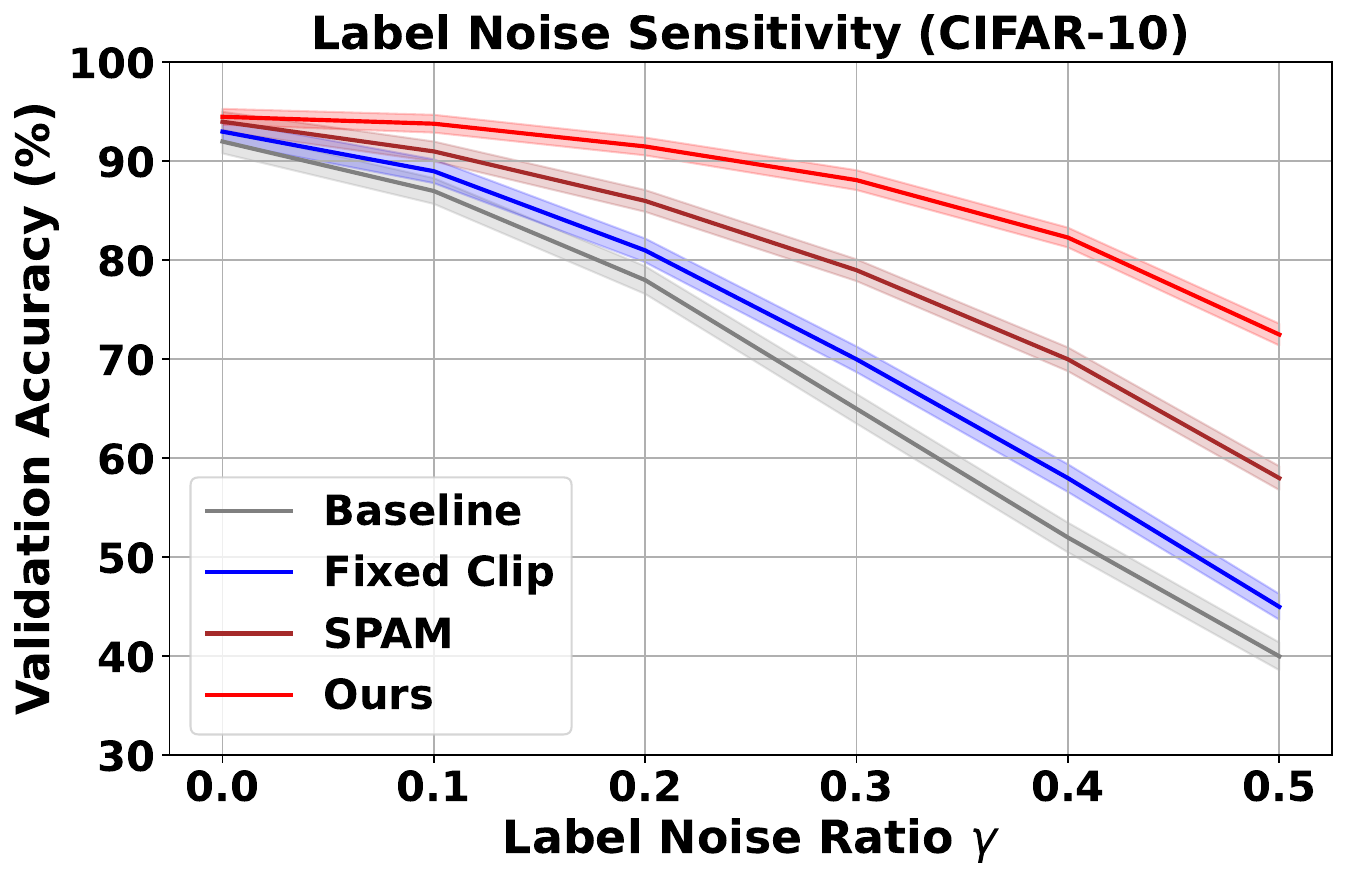}
        \caption{Accuracy under increasing label noise.}
        \label{fig:label_noise}
    \end{subfigure}
    \hfill
    \begin{subfigure}[t]{0.48\linewidth}
        \centering
        \includegraphics[width=\linewidth]{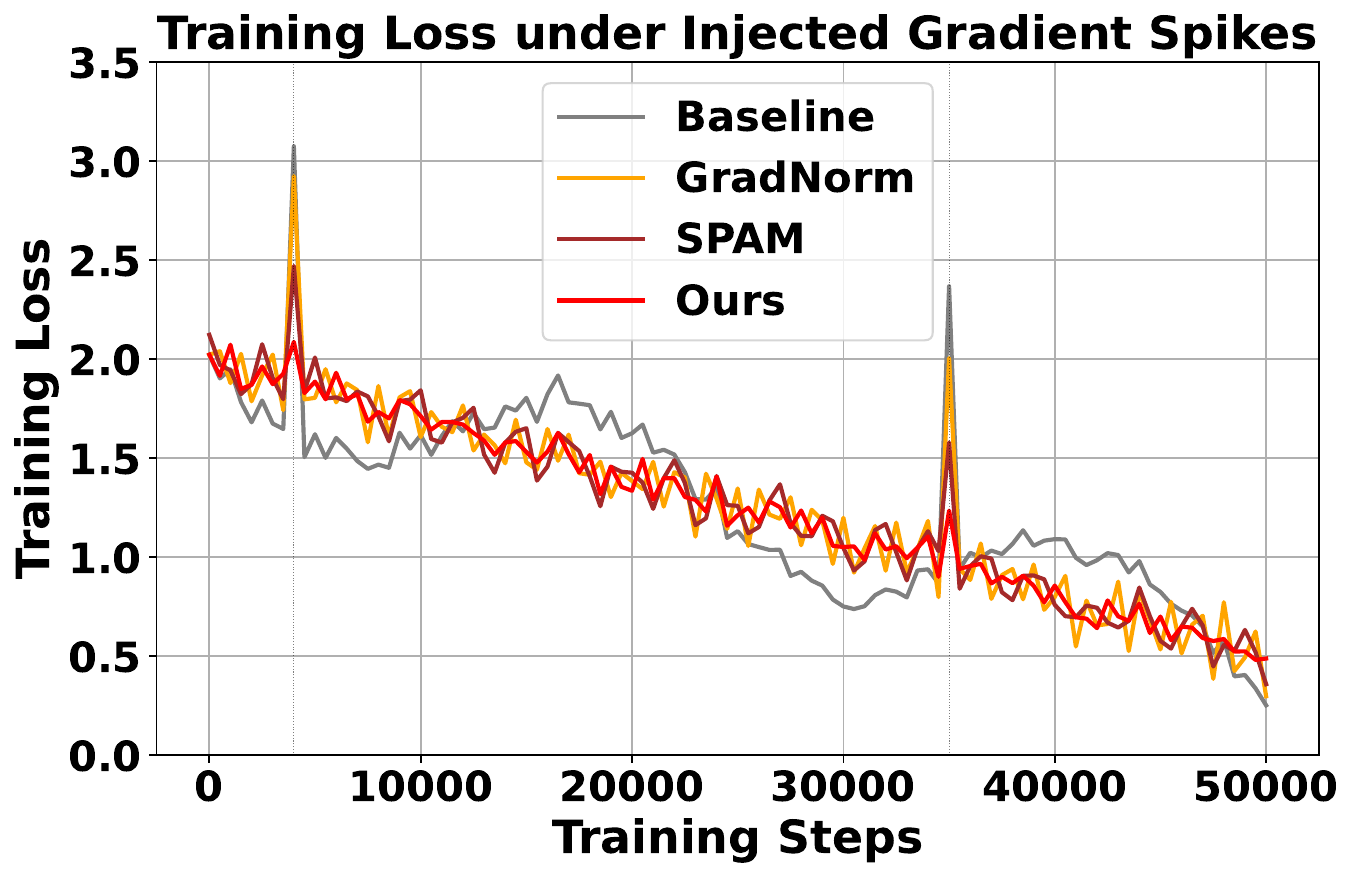}
        \caption{Training loss with injected gradient spikes.}
        \label{fig:spike_loss}
    \end{subfigure}
    \caption{Robustness under label noise and gradient spikes.}
    \label{fig:robustness_fig}
\end{figure}

We test the robustness of different optimization methods under three types of training-time perturbations: label noise, gradient spikes, and batch size variation. For label noise, we randomly corrupt a fraction $\gamma$ of CIFAR-10 labels. As shown in Figure~\ref{fig:robustness_fig}~(\subref{fig:label_noise}), our method maintains high accuracy up to $\gamma = 40\%$, while baselines drop significantly beyond 20\%. For gradient spikes, we inject 5$\times$ scaled gradients at 2\% of steps. Figure~\ref{fig:robustness_fig}~(\subref{fig:spike_loss}) shows that our method recovers quickly and avoids oscillation, unlike fixed clipping or GradNorm.

To simulate batch-induced variance, we alternate small (B=16) and large (B=512) batch sizes every 1k steps. Table~\ref{tab:batch_var} reports the standard deviation and maximum of $\eta_t \|g_t\|$ in different intervals \cite{you2025mover}. Our method exhibits the lowest volatility and worst-case magnitudes across the board. These results demonstrate that our shaping strategy consistently stabilizes training under various perturbations, thanks to its continuous and adaptive design.

\begin{table}[ht]
    \centering
    \small
    \setlength{\tabcolsep}{4pt}
    \renewcommand{\arraystretch}{0.98}
    \begin{tabular}{lcccccc}
    \toprule
    \multirow{2}{*}{\textbf{Method}} & \multicolumn{3}{c}{\textbf{Std. of $\eta_t \|g_t\|$}} & \multicolumn{3}{c}{\textbf{Max $\eta_t \|g_t\|$}} \\
    \cmidrule(lr){2-4} \cmidrule(lr){5-7}
    & 0-10k & 10k-30k & 30k-50k & 0-10k & 10k-30k & 30k-50k \\
    \midrule
    Baseline         & 0.420 & 0.388 & 0.341 & 3.84 & 3.51 & 2.94 \\
    Fixed Clipping   & 0.297 & 0.263 & 0.224 & 2.78 & 2.33 & 1.94 \\
    GradNorm         & 0.244 & 0.218 & 0.205 & 2.45 & 2.10 & 1.78 \\
    ZClip            & 0.199 & 0.183 & 0.168 & 2.11 & 1.93 & 1.60 \\
    \textbf{Ours}    & \textbf{0.131} & \textbf{0.112} & \textbf{0.097} & \textbf{1.73} & \textbf{1.58} & \textbf{1.42} \\
    \bottomrule
    \end{tabular}
    \caption{Std. deviation and max of update magnitudes under batch size shifts. Lower values imply stronger robustness.}
    \label{tab:batch_var}
\end{table}

\subsection{Ablation Study}

We assess the contribution of each component in our method by removing or modifying submodules. Metrics include final validation accuracy, early-stage stability (variance of $\eta_t \|g_t\|$ in the first 10k steps), and average update magnitude. The following components are ablated individually:
\begin{itemize}
    \item \textbf{Dynamic $\tau$}: Replaced with fixed $\tau = 1$
    \item \textbf{Power shaping ($\alpha$)}: Replaced with hard clipping
    \item \textbf{EMA smoothing}: Removed exponential averaging
    \item \textbf{Per-layer adaptivity}: Replaced with global shaping
\end{itemize}

\begin{figure}[htbp]
    \centering
    \begin{subfigure}[t]{0.46\linewidth}
        \centering
        \includegraphics[width=\linewidth]{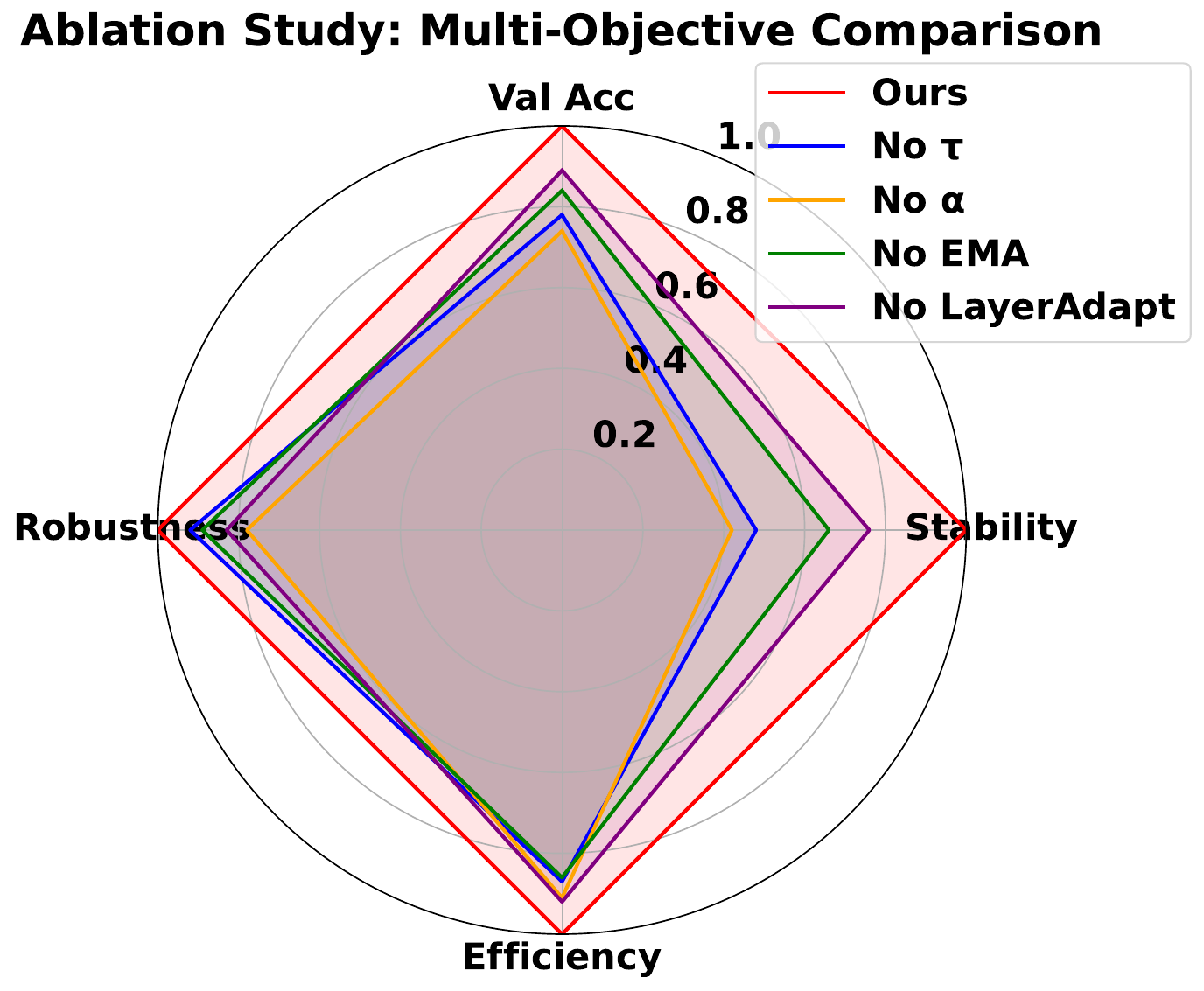}
        \caption{Radar plot: accuracy, stability, efficiency, robustness.}
        \label{fig:radar_ablation}
    \end{subfigure}
    \hfill
    \begin{subfigure}[t]{0.50\linewidth}
        \centering
        \includegraphics[width=\linewidth]{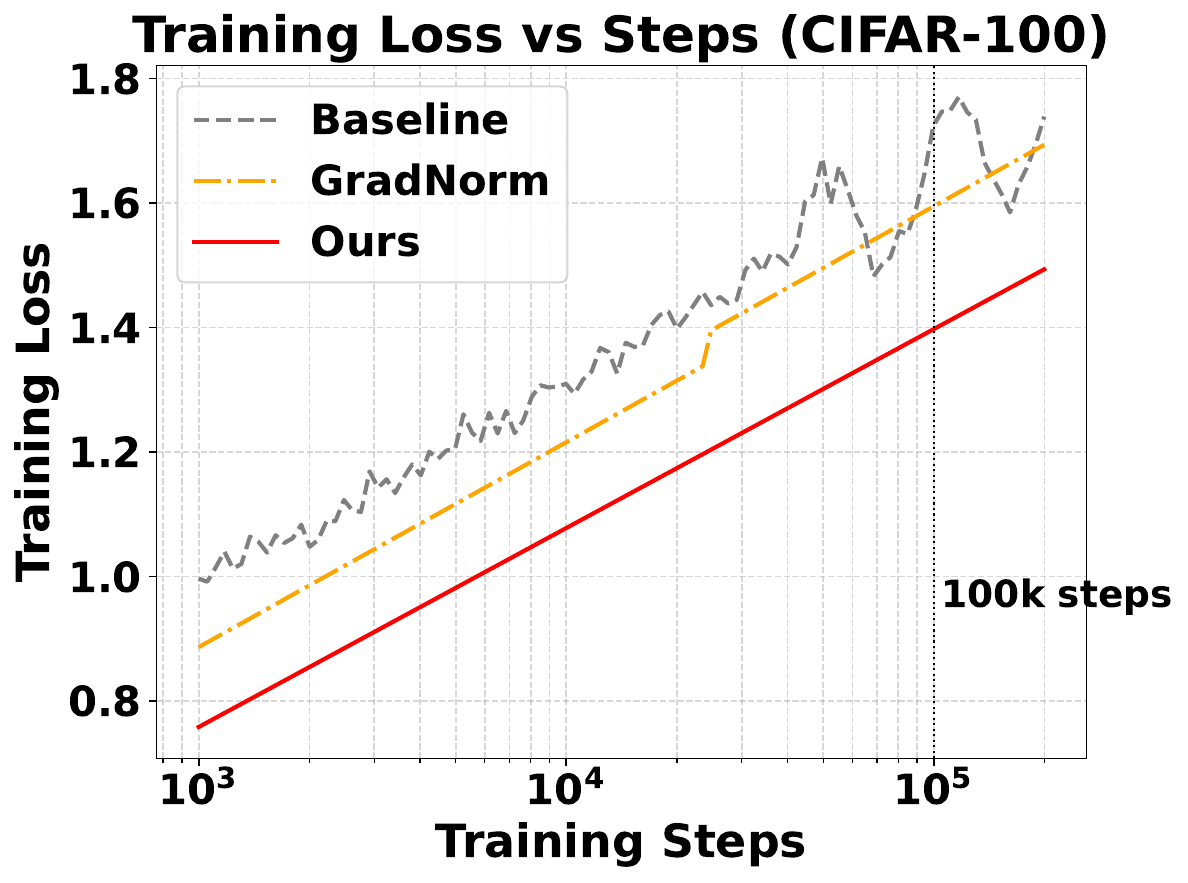}
        \caption{Training loss vs. steps (CIFAR-100).}
        \label{fig:long_training}
    \end{subfigure}
    \caption{Ablation and scaling analysis. Our method maintains robustness and performance across conditions.}
    \label{fig:visual_ablation_scaling}
\end{figure}

Table~\ref{tab:ablation_table} and Figure~\ref{fig:radar_ablation} show that each component contributes to stability or accuracy. The largest performance drop occurs when removing dynamic thresholds or shaping, highlighting the importance of smooth, adaptive modulation \cite{you2024application}. The full configuration yields the best trade-off across metrics.

\begin{table}[ht]
    \centering
    \setlength{\tabcolsep}{3pt}
    \begin{tabular}{lccc}
        \toprule
        \textbf{Configuration} & \textbf{Val Acc (\%)} & \textbf{Early Var} & \textbf{Avg $\eta_t \|g_t\|$} \\
        \midrule
        Full (Ours)               & \textbf{91.3} & \textbf{0.011} & \textbf{1.14} \\
        w/o Dynamic $\tau$        & 89.1 & 0.028 & 1.33 \\
        w/o Power Shaping ($\alpha$) & 88.7 & 0.034 & 1.26 \\
        w/o EMA                   & 89.5 & 0.021 & 1.22 \\
        w/o Per-layer Adaptivity  & 90.1 & 0.018 & 1.17 \\
        \bottomrule
    \end{tabular}
    \caption{Ablation study on CIFAR-10. Each component contributes to either stability or performance.}
    \label{tab:ablation_table}
\end{table}

\subsection{Scaling Behavior}

We evaluate whether our method generalizes to larger models and longer training runs. On ImageNet, we test ResNet-18, ResNet-50, and ViT-Tiny. As shown in Table~\ref{tab:scaling_model}, our approach consistently improves top-1 accuracy across model scales, while fixed clipping and GradNorm tend to plateau as complexity grows.

\begin{table}[ht]
    \centering
    \begin{tabular}{lccc}
    \toprule
    \textbf{Method} & ResNet-18 & ResNet-50 & ViT-Tiny \\
    \midrule
    Baseline         & 70.4 & 74.1 & 73.3 \\
    Fixed Clipping   & 72.6 & 75.3 & 74.2 \\
    GradNorm         & 73.1 & 75.5 & 74.6 \\
    \textbf{Ours}    & \textbf{74.4} & \textbf{76.6} & \textbf{75.9} \\
    \bottomrule
    \end{tabular}
    \caption{Top-1 validation accuracy (\%) across model scales.}
    \label{tab:scaling_model}
\end{table}

We also examine long-horizon behavior by training on CIFAR-100 for up to 200k steps. As shown in Figure~\ref{fig:long_training}, our method sustains smooth, stable loss descent throughout. In contrast, baseline methods show late-stage oscillation or stagnation. These results suggest that our shaping strategy remains effective across scale-without requiring retuning for larger models or extended training schedules.

\section{Conclusion}

We proposed \textbf{SPAMP}, a unified framework that reframes gradient clipping as a smooth, adaptive shaping process grounded in per-layer statistics. By highlighting the central role of update magnitude $\eta_t \|g_t\|$, we connected clipping, warmup, and gradient scaling under a functional perspective. Our method improves training stability and convergence across architectures, offering a principled alternative to rigid thresholding. This work opens the door to more flexible, learnable forms of update modulation in large-scale optimization.

\bibliographystyle{ACM-Reference-Format}
\bibliography{sample-base}

\end{document}